\documentclass{bmvc2k}

\title{Efficient Image Super-Resolution via\\Symmetric Visual Attention Network}

\usepackage{marvosym}
\addauthor{Chengxu Wu}{woox929@163.com}{1}
\addauthor{Qinrui Fan}{fanqinr@gmail.com}{1}
\addauthor{Shu Hu}{hu968@purdue.edu}{2}
\addauthor{Xi Wu}{xi.wu@cuit.edu.cn}{1}
\addauthor{Xin Wang}{xwang56@albany.edu}{3, \Letter}
\addauthor{Jing Hu}{jing_hu09@163.com}{1, \Letter}

\addinstitution{
 Chengdu University of Information Technology\\
 Chengdu, China
}
\addinstitution{
Purdue University in Indianapolis, \\
IN, USA}
\addinstitution{
University at Albany, SUNY, \\
New York, USA}

\runninghead{C.Wu, Fan, S.Hu, X.Wu, Wang, J.Hu}{Efficient Image SR via SVAN}

\def\etal{\emph{et al}\bmvaOneDot}

\usepackage{multirow}
\begin{document}
\maketitle
\begin{abstract}
An important development direction in the Single-Image Super-Resolution (SISR) algorithms is to improve the efficiency of the algorithms. Recently, efficient Super-Resolution (SR) research focuses on reducing model complexity and improving efficiency through improved deep small kernel convolution, leading to a small receptive field. The large receptive field obtained by large kernel convolution can significantly improve image quality, but the computational cost is too high. To improve the reconstruction details of efficient super-resolution reconstruction, we propose a Symmetric Visual Attention Network (SVAN) by applying large receptive fields. The SVAN decomposes a large kernel convolution into three different combinations of convolution operations and combines them with an attention mechanism to form a Symmetric Large Kernel Attention Block (SLKAB), which forms a symmetric attention block with a bottleneck structure by the size of the receptive field in the convolution combination to extract depth features effectively as the basic component of the SVAN. Our network gets a large receptive field while minimizing the number of parameters and improving the perceptual ability of the model. The experimental results show that the proposed SVAN can obtain high-quality super-resolution reconstruction results using only about 30\% of the parameters of existing SOTA methods.

\end{abstract}

\section{Introduction}
\label{Introduction}

Single Image Super Resolution (SISR) is the process of recovering a High-Resolution (HR) image from a single Low-Resolution (LR) image that has undergone a degradation process. The strong demand for high-resolution images is the result of rapidly evolving image processing techniques \cite{zamir2022restormer} and popular computer vision applications \cite{gu2022multi}. SISR techniques provide both better visual fidelity and enhanced detail of image information and are widely used in various computer vision tasks \cite{dai2016image} and real-world scenarios \cite{li2021review,sun2021learning,dong2020remote}.

With the great success of deep learning technology in the field of computer vision, SISR algorithms based on deep learning have gained widespread attention and rapid development after SRCNN \cite{dong2016accelerating} pioneered the combination of deep learning and SISR. However, to improve the performance of SR, existing models \cite{liang2021swinir,chen2022activating} often use complex deep network structures, which means that advanced work requires high computational costs to cope with the huge number of parameters, hindering the adoption and deployment of SR models \cite{zhang2021edge}. Therefore, it is crucial to achieve a balance between image quality and the number of parameters for efficient models. 

\begin{figure*}[t]
\begin{center}
\includegraphics[width=0.8\textwidth]{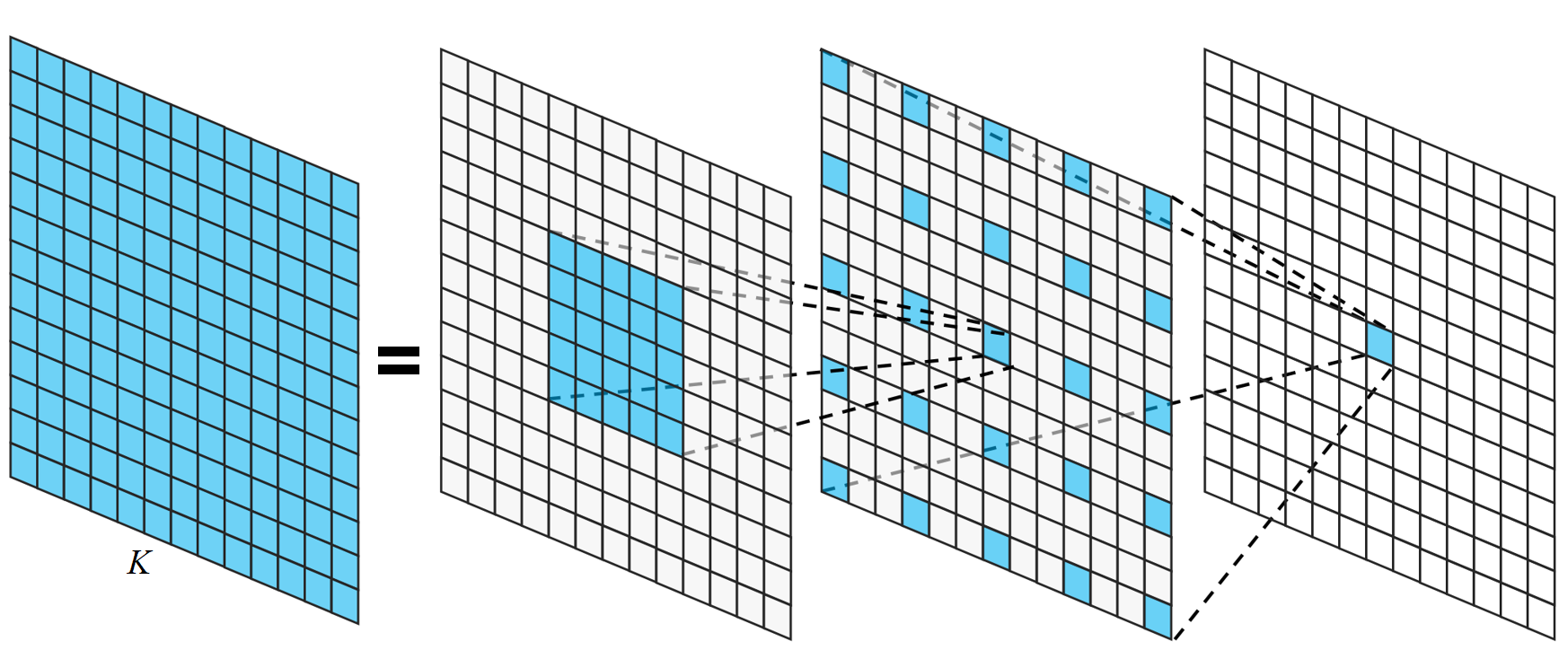}
\end{center}
\caption{Large kernel convolution with kernel 13 can be decomposed into a 5$\times$5 depth-wise convolution and a 5$\times$5 depth-wise dilation convolution with a dilation of 3. The figure shows the convolution combination used in our model: a 5$\times$5 depth-wise convolution and a 5$\times$5 depth-wise dilation convolution with a dilation of 3, and a 1$\times$1 point convolution. The blue color shows the kernel. Note: there are zero paddings in the figure.}
\label{fig:conv}
\end{figure*}

To improve model efficiency and reduce the model size, researchers have proposed various methods to reduce the complexity of the models, including efficient operation design \cite{kong2022residual, li2022ntire,luo2023fast}, neural structure search \cite{chu2021fast,guo2020hierarchical}, knowledge distillation \cite{gao2019image,hui2019lightweight}, and structural parametric reconstruction \cite{zhang2021edge}.
The above methods are mainly based on improved deep small kernel convolution \cite{hui2019lightweight, li2022ntire, zhang2021edge,luo2022deep}. The receptive field obtained by stacking network depth is very small, but Shamsolmoali \etal~\cite{shamsolmoali2019single} concluded that the receptive field has a more significant etimpact on image quality than the network depth. A large receptive field can capture more global feature information, which can improve the fineness of the reconstruction results in the SR task of pixel-by-pixel prediction. The size of the receptive field is proportional to the size of the convolution kernel \cite{ding2022scaling}, so using a large kernel convolution is an intuitive but weighty way to obtain an effective receptive field. To reduce the computational cost of large kernel convolution, using depth-wise convolution and depth-wise dilation convolution are effective alternatives \cite{guo2022visual}. As shown in the Figure \ref{fig:conv} a large kernel convolution can be decomposed to a depth-wise convolution in a local space and a depth-wise dilation convolution in a long-range space. Unlike traditional convolution, depth-wise convolution only computes the feature map at the spatial level and cannot expand the dimensions, which will reduce the information interaction between different feature maps, so it is challenging to capture enough interaction information of channels and spatial. Therefore, we use a combination of three lightweight convolution operations to expand the receptive field and design the arrangement structure to improve learning ability in our efficient network design.

In this paper, we develop a simple but effective SR method called Symmetric Visual Attention Network (SVAN), whose core idea is to improve the reconstruction quality of SR images by using a large receptive field. By combining three convolution operations to obtain a lightweight and efficient large kernel attention block, the receptive field of the network is expanded to enhance efficient SR performance while effectively controlling the number of parameters. Specifically, a spatial 5$\times$5 depth-wise convolution, a spatial 5$\times$5 depth-wise dilation convolution with a dilation of 3, and a channel point convolution are combined to achieve the same receptive field of a large kernel convolution with kernel 17 with much fewer number of parameters, and can better fuse spatial and channel information. The combination of convolution forms an attention block with a large receptive field, which enhances local contextual information extraction and improves the interaction of spatial and channel dimensional information. Next, two sets of attention blocks are symmetrically arranged to obtain symmetric large kernel attention blocks (SLKAB), which forms bottleneck structured attention according to the size of different convolution layer receptive field. And the bottleneck structure \cite{zhou2020rethinking} can effectively fuse multi-scale information while enhancing the model's global information and local information perception using different receptive field sizes, which can further compress and refine the extracted features and improve the learning and expression ability of the network. Using symmetric arrangement \cite{gao2022lightweight} can improve the expressiveness, generalization ability, and computational efficiency of the network. Therefore, the symmetric structure and the bottleneck design of the receptive field size are introduced into the attention module of our network.

The contributions of this paper are three-fold: 
\begin{enumerate}
\item Our SVAN improves model efficiency by constructing convolution combinations to form a large kernel attention with a large receptive field, which has fewer parameters compared to existing efficient SR methods. It leads to a lightweight large kernel SR model that enables direct and effective expansion of the network receptive field.

\item The proposed SLKAB enhances the extraction of depth features with bottleneck receptive field structure and symmetric attention structure, which further improves the learning ability of the network.

\item From our experiments, our SVAN shows better results than the existing state-of-the-art methods in terms of both parameter number and FLOPs while maintaining high image quality.
\end{enumerate}

The rest of this paper is organized as follows. Related work is described in Section \ref{Related Work}, and SVAN and SLKAB of our model are described in Section \ref{Method}. Section \ref{Experiment} shows extensive experiments on the performance evaluation of our proposed SVAN. Finally, conclusions are drawn in Section \ref{Conclusions}.

\section{Related Work}
\label{Related Work}

\subsection{Efficient Super-Resolution}

Efficient SR networks are designed to reduce model complexity and computational cost. For this purpose, DRCN \cite{kim2016deeply} introduces a deep recursive convolution network to reduce the number of parameters, which is too deep and difficult to train. CARN \cite{ahn2018fast} proposes an efficient cascaded residual network using group convolution and recurrent networks to eliminate redundant parameters, but the model has a long inference time. After IDN \cite{hui2018fast} proposes a residual feature distillation structure, IMDN \cite{hui2019lightweight} uses a channel splitting strategy to improve IDN by information multi-distillation block and proposes a lightweight information multi-distillation network, but with a large parameter number. RFDN \cite{liu2020residual}, on the other hand, uses feature distillation connection instead of information distillation and proposes a residual feature distillation network, but with a slower inference speed. Existing studies mainly use various complex inter-layer connections and improved small kernel convolution to improve SR efficiency, but the receptive field of the network is small and the reconstruction details need to be improved.

\subsection{Large Kernel for Attention}

The attention mechanism can help SR models to accurately focus on important details in images and improve the quality of reconstructed images. There are three types of attention: channel attention, spatial attention, and self-attention. RCAN \cite{zhang2018image} proposes a deep residual channel attention network to adaptively readjust the interdependence between channels by channel attention mechanism. HAN \cite{niu2020single} proposes a layer attention module and a channel-space attention module to model the information features between layers, channels, and locations. HAT \cite{chen2022activating} further proposes various hybrid attention schemes that combine channel attention and self-attention. These works demonstrate the prominent role of the attention mechanism in SR.

With the development of efficient convolution techniques, large kernel convolution has recently gained a lot of attention \cite{liu2021swin, liang2021swinir}. Large convolution kernels have a larger receptive field to obtain more global information, and recent studies have demonstrated that large kernel convolution has better performance in attention networks. ConvNeXt \cite{liu2022convnet} redesigned a standard ResNet using 7$\times$7 kernels and obtained comparable results to Transformer \cite{vaswani2017attention}. RepLKNet \cite{ding2022scaling} builds a pure CNN model with 31$\times$31 kernel convolutions, outperforming SOTA Transformer-based methods. VAN \cite{guo2022visual} analyzes visual attention and proposes large kernel attention based on deep convolution. These methods of applying large kernel convolution models to solve vision tasks provide a reference for our research on the use of large kernel convolution in efficient SR.

\section{Method}
\label{Method}

This section first details the overall pipeline of our proposed Symmetric Visual Attention Network (SVAN). We further elaborated on the Symmetric Large Kernel Attention Block (SLKAB) which is the basic module of SVAN.

\begin{figure*}[t]
\begin{center}
\includegraphics[width=0.95\textwidth]{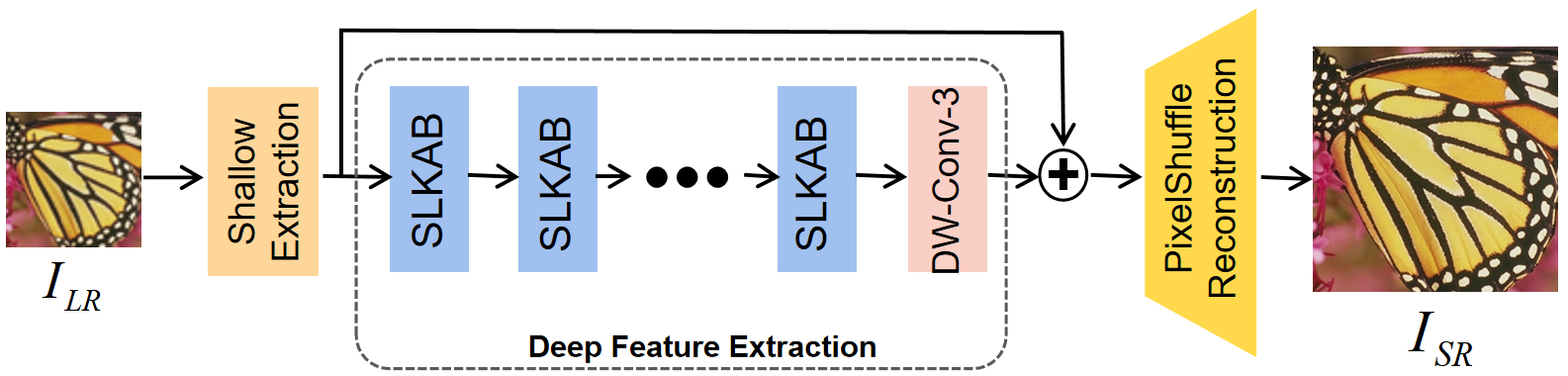}
\end{center}
   \caption{The architecture of Symmetric Visual Attention Network. SVAN contains three main parts: shallow feature extraction module, deep feature extraction module, and pixel shuffle reconstruction module.}
\label{fig:svan}
\end{figure*}

\subsection{Symmetric Visual Attention Network}

The overall structure of the lightweight Symmetric Visual Attention Network (SVAN) is shown in Figure \ref{fig:svan}, which contains three main parts: shallow feature extraction module, deep feature extraction module, and pixel shuffle reconstruction module.

We denote the {$I_{LR}$} and {$I_{SR}$} as the input and output of the SVAN. First, we use a single 3$\times$3 convolution layer to extract shallow features.
\begin{equation}
    x_{0}=f_{ext}(I_{LR})
\end{equation}
where {$f_{ext}(\cdot)$} is the convolution operation for shallow feature extraction and {$x_{0}$} is the extracted feature map. Then, we use multiple SLKAB blocks for depth feature extraction. This process can be expressed as follows:
\begin{equation}
    x_{n}=f_{SLKAB}^{n}(f_{SLKAB}^{n-1}(\cdots f_{SLKAB}^{0}(x_{0})))
\end{equation}
where {$f_{SLKAB}^{n}(\cdot)$} denotes the n-th SLKAB function and {$x_{n}$} denotes the feature map of the n-th SLKAB output. At the end of the deep feature extraction stage, we use {$f_{ref}(\cdot)$}, a 3$\times$3 depth-wise dilation convolution with a dilation of 3, to further reduce the number of parameters while refining the deep feature map with residual concatenation to {$x_{0}$} :
\begin{equation}
    x_{map}=f_{ref}(x_{n}) + x_{0}
\end{equation}
Finally, the features are upsampled using the reconstruction module to reach the HR size.
\begin{equation}
    I_{SR}=f_{rec}(x_{map})
\end{equation}
{$f_{rec}(\cdot)$} denotes the reconstruction module consisting of a single 3$\times$3 convolution layer and a pixel-shuffle \cite{shi2016real} layer, and {$I_{SR}$} is the final result of the network.

\subsection{Symmetric Large Kernel Attention Block }

\begin{figure*}[t]
\begin{center}
\includegraphics[width=0.95\textwidth]{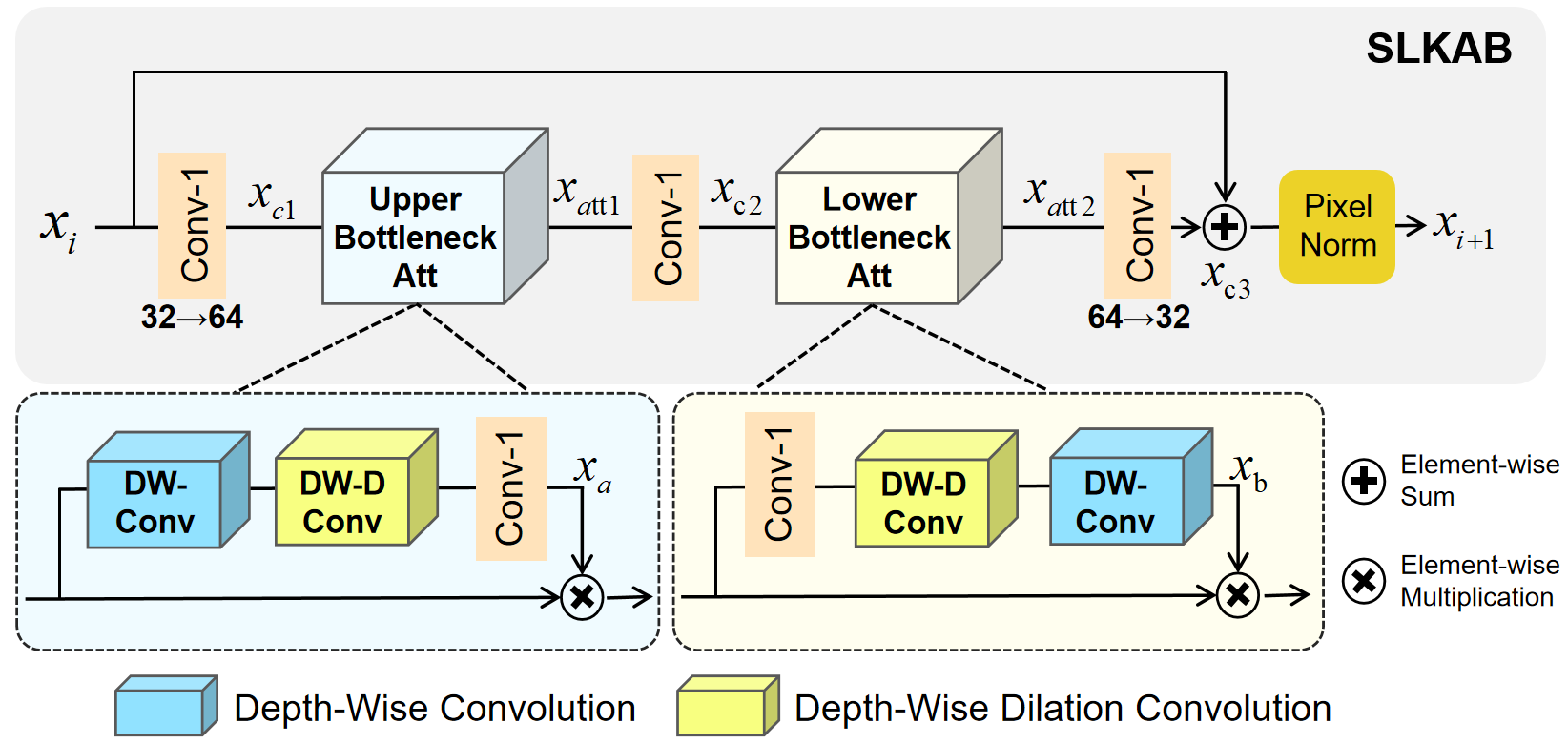}
\end{center}
   \caption{The architecture of Symmetric Large Kernel Attention Block. We perform a receptive field size bottleneck structure and symmetrical design for attention in SLKAB.}
\label{fig:slkab}
\end{figure*}

SLKAB utilizes a 5$\times$5 depth-wise convolution, a 5$\times$5 depth-wise dilation convolution with a dilation of 3, and a 1$\times$1 point convolution to reach the receptive field of a large kernel convolution using 17$\times$17. Such a combination of convolution takes into account both spatial and channel information, and also greatly compresses the number of parameters. 

We use two sets of convolution combinations symmetrically arranged in the block design to form a dual attention module. By forming a symmetric attention structure with a bottleneck structure through large-small-small-large size receptive fields, the feature information is extracted interactively to enhance the generalization ability of the module and balance the parameters and performance. When the original feature map is input to upper bottleneck attention, the feature extraction is first performed by the convolution layer in the large receptive field to ensure the maximum information of the original input. The point convolution layer in the small receptive field performs feature refinement and is then input into lower bottleneck attention. The lower bottleneck attention has the opposite arrangement of receptive field sizes so that a symmetrical attention structure of the bottleneck is formed.

As shown in Figure \ref{fig:slkab}. as the input of the n-th block {$x_{i}$} is first expanded from 32 to 64 channels using 1$\times$1 convolution {$conv(\cdot)$} and GELU \cite{hendrycks2016gaussian} activation {$gelu(\cdot)$} to obtain more information.
\begin{equation}
    x_{c1}=gelu(conv(x_{i}))
\end{equation}
The features generated by the attention branch are fused with the original features {$x_{c1}$} using elemental multiplication. {$DW(\cdot)$} and {$DWD(\cdot)$} are depth-wise convolution and depth-wise dilation convolution.
\begin{equation}
    x_{att1}=UAB(DW(DWD(conv(x_{c1}))) \bigotimes x_{c1})
\end{equation}
{$UAB(\cdot)$} denotes the upper bottleneck attention, {$x_{att1}$} after the second 1$\times$1 convolution {$conv(\cdot)$} gets {$x_{c2}$} as input to the lower bottleneck attention, and similarly, we get:
\begin{equation}
    x_{att2}=LAB(conv(DWD(DW(x_{c1}))) \bigotimes x_{c2})
\end{equation}
After the lower bottleneck attention {$LAB(\cdot)$}, the third 1$\times$1 convolution layer adjusts the number of channels back to 32 and fuses the original input by skip the connection:
\begin{equation}
    x_{c3}=conv(x_{att2}) + x_{i}
\end{equation}
Finally, pixel normalization {$pn(\cdot)$} is used to improve the stability of the training and to obtain the output {$x_{i+1}$} of SLKAB.
\begin{equation}
    x_{i+1}=pn(x_{c3})
\end{equation}

\section{Experiment}
\label{Experiment}

\subsection{Dataset and implementation details}

{\bf Dataset and evaluation metrics.} The training set consisted of 2650 images from Flickr2K \cite{lim2017enhanced} and 800 images from DIV2K \cite{agustsson2017ntire}. Our models were evaluated on widely used benchmark datasets: Set5 \cite{bevilacqua2012low}, Set14 \cite{zeyde2012single}, BSD100 \cite{martin2001database}, and Urban100 \cite{huang2015single}. We train our model on RGB channels and augment data with random rotations and flipping. We calculated the PSNR and SSIM \cite{wang2004image} on the Y channel in the YCbCr space as quantitative measurements.

{\bf Implementation details.} Our implementation of SVAN contains 32 channels and 7 SLKAB blocks. To better extract deep information, the number of channels in SLKAB is expanded to 64, and the adjustment of the number of channels is achieved by 1$\times$1 convolution. The channel settings of the convolution and pixel-shuffle layers in the reconstruction module adjust according to scale factors.


During the training process, The patch of size 64 is random cropping from LR images as input, the minibatch size is set to 64, the Adam optimizer \cite{kingma2014adam} is used for optimization, and the training is divided into two stages.

In the first stage, a pre-training of 2000 epochs is performed using the minimized L1 loss function, and the learning rate is set to $1\times 10^{-3}$ and halved every 500 epochs.

In the second stage, load the pre-trained model from the first stage and using the minimized L1 loss, the initial learning rate is set to $1\times 10^{-4}$, the learning rate is adjusted by cosine annealing with a period of 20, and the input patch size is set to 64 and 128 for each training once each of 3000 epochs. Finally fine-tuning of 3000 epochs using L2 loss, with an initial learning rate set of $5\times 10^{-4}$, halving every 300 epochs.

\subsection{Comparison with competitive methods}

\begin{table}[ht]
\begin{center}
\scalebox{0.76}{
\begin{tabular}{|c|l|cccccc|}
\hline
Scale & Method & Params[K] & FLOPs[G] & Set5 & Set14 & BSD100 & Urban100\\ \hline\hline

\multirow{8}{*}{$\times$2} & Bicubic & - & - & 33.66/0.9299 & 30.24/0.8688 & 29.56/0.8431 & 26.88/0.8403\\ 

\multirow{8}{*}{ } & SRCNN \cite{dong2016accelerating} & 57 & 3.8 & 36.66/0.9542 & 32.45/0.9067 & 31.36/0.8879 & 29.50/0.8946\\

\multirow{8}{*}{ } & CARN \cite{ahn2018fast} & 1592 & 63.8 & 36.66/0.9542 & 32.45/0.9067 & 31.36/0.8879 & 29.50/0.8946\\

\multirow{8}{*}{ } & IMDN \cite{hui2019lightweight} & 694 & 57.2 & 38.00/0.9605 & 33.63/0.9177 & 32.19/0.8996 & 32.17/0.9283\\

\multirow{8}{*}{ } & RFDN \cite{liu2020residual} & 534 & 35.2 & 38.05/0.9606 & 33.68/0.9184 & 32.16/0.8994 & 32.12/0.9278\\

\multirow{8}{*}{ } & ECBSR \cite{zhang2021edge} & 596 & 25.6 & 37.90/0.9615 & 33.34/0.9178 & 32.10/0.9018 & 31.71/0.9250\\

\multirow{8}{*}{ } & RLFN \cite{kong2022residual} & 527 & 32.9 & 38.07/0.9607 & 33.72/0.9187 & 32.22/0.9000 & 32.33/0.9299\\

\multirow{8}{*}{ } & SVAN(ours) & {\color[HTML]{FE0000} 173} & {\color[HTML]{FE0000} 11.0} & {\bf 37.70}/{\bf 0.9592} & {\bf 33.40}/{\bf 0.9158} & {\bf 31.98}/{\bf 0.8964} & {\bf 31.44}/{\bf 0.9220}\\
\hline\hline

\multirow{6}{*}{$\times$3} & Bicubic & - & - & 30.39/0.8682 & 27.55/0.7742 & 27.21/0.7385 & 24.46/0.7349\\ 

\multirow{6}{*}{ } & SRCNN \cite{dong2016accelerating} & 57 & 3.8 & 32.75/0.9090 & 29.30/0.8215 & 28.41/0.7863 & 26.24/0.7989\\

\multirow{6}{*}{ } & CARN \cite{ahn2018fast} & 1592 & 76.2 & 34.29/0.9255 & 30.29/0.8407 & 29.06/0.8034 & 28.06/0.8493\\

\multirow{6}{*}{ } & IMDN \cite{hui2019lightweight} & 703 & 57.7 & 34.36/0.9270 &  30.32/0.8417 & 29.09/0.8046 & 28.17/0.8519\\

\multirow{6}{*}{ } & RFDN \cite{liu2020residual} & 541 & 35.6 & 34.41/0.9273 & 30.34/0.8420 & 29.09/0.8050 & 28.21/0.8525\\

\multirow{6}{*}{ } & SVAN(ours) & {\color[HTML]{FE0000} 177} & {\color[HTML]{FE0000} 11.3} & {\bf 33.92}/{\bf 0.9228} & {\bf 30.12}/{\bf 0.8372} & {\bf 28.91}/{\bf 0.7987} & {\bf 27.52}/{\bf 0.8388}\\
\hline\hline

\multirow{8}{*}{$\times$4} & Bicubic & - & - & 28.42/0.8104 & 26.00/0.7027 & 25.96/0.6675 & 23.14/0.6577\\

\multirow{8}{*}{ } & SRCNN \cite{dong2016accelerating} & 57 & 3.8 & 30.48/0.8626 & 27.50/0.7513 & 26.90/0.7101 & 24.52/0.7221\\

\multirow{8}{*}{ } & CARN \cite{ahn2018fast} & 1592 & 104.5 & 32.13/0.8937 & 28.60/0.7806 & 27.58/0.7349 & 26.07/0.7837\\

\multirow{8}{*}{ } & IMDN \cite{hui2019lightweight} & 715 & 58.5 & 32.21/0.8948 & 28.58/0.7811 & 27.56/0.7353 & 26.04/0.7838\\

\multirow{8}{*}{ } & RFDN \cite{liu2020residual} & 550 & 36.2 & 32.24/0.8952 & 28.61/0.7819 & 27.57/0.7360 & 26.11/0.7858\\

\multirow{8}{*}{ } & ECBSR \cite{zhang2021edge} & 603 & 28.3 & 31.92/0.8946 & 28.34/0.7817 & 27.48/0.7393 & 25.81/0.7773\\

\multirow{8}{*}{ } & RLFN \cite{kong2022residual} & 527 & 34.0 & 32.24/0.8952 & 28.62/0.7813 & 27.60/0.7364 & 26.17/0.7877\\

\multirow{8}{*}{ } & SVAN(ours) & {\color[HTML]{FE0000} 183} & {\color[HTML]{FE0000} 11.7} & {\bf 31.76}/{\bf 0.8890} & {\bf 28.30}/{\bf 0.7736} & {\bf 27.41}/{\bf 0.7285} & {\bf 25.56}/{\bf 0.7685}\\

\hline
\end{tabular}}
\end{center}
\caption{Quantitative results of the SOTA models on four benchmark datasets. Params and FLOPs are the total numbers of network parameters and floating-point operations. The FLOPs calculation corresponds to images of a size of 256$\times$256. Our model's Params and FLOPs results are highlighted in {\color[HTML]{FE0000} Red}, and the PSNR/SSIM results are highlighted in {\bf bold}.}
\label{table1}
\end{table}

{\bf Quantitative evaluations.} We compared the proposed SVAN with existing common efficient SR models with scale factors of $\times$2, $\times$3, and $\times$4, including SRCNN \cite{dong2016accelerating}, CARN \cite{ahn2018fast}, IMDN \cite{hui2019lightweight}, RFDN \cite{liu2020residual}, ECBSR \cite{zhang2021edge}, and RLFN \cite{kong2022residual}. 

Quantitative performance comparisons on several benchmark datasets are shown in Table \ref{table1}. We also list the number of parameters and FLOPs. Compared with other SOTA models, our SVAN achieves high-quality SR reconstructions with extremely few parameters and FLOPs, with only a slight performance loss in PSNR and SSIM. Specifically, our SVAN $\times$4 uses 34.72\% and 33.27\% of the number of parameters of RLFN $\times$4 and RFDN $\times$4, with an average performance decrease of only 0.38 dB in PSNR on the four test datasets. We can conclude that our model can still obtain competitive results with SOTA models after a significant reduction in the number of parameters, leading to lightweight and efficient models, showing that SVAN is the right direction to explore for efficient SR.

\begin{figure*}[t]
\begin{center}
\includegraphics[width=1\textwidth]{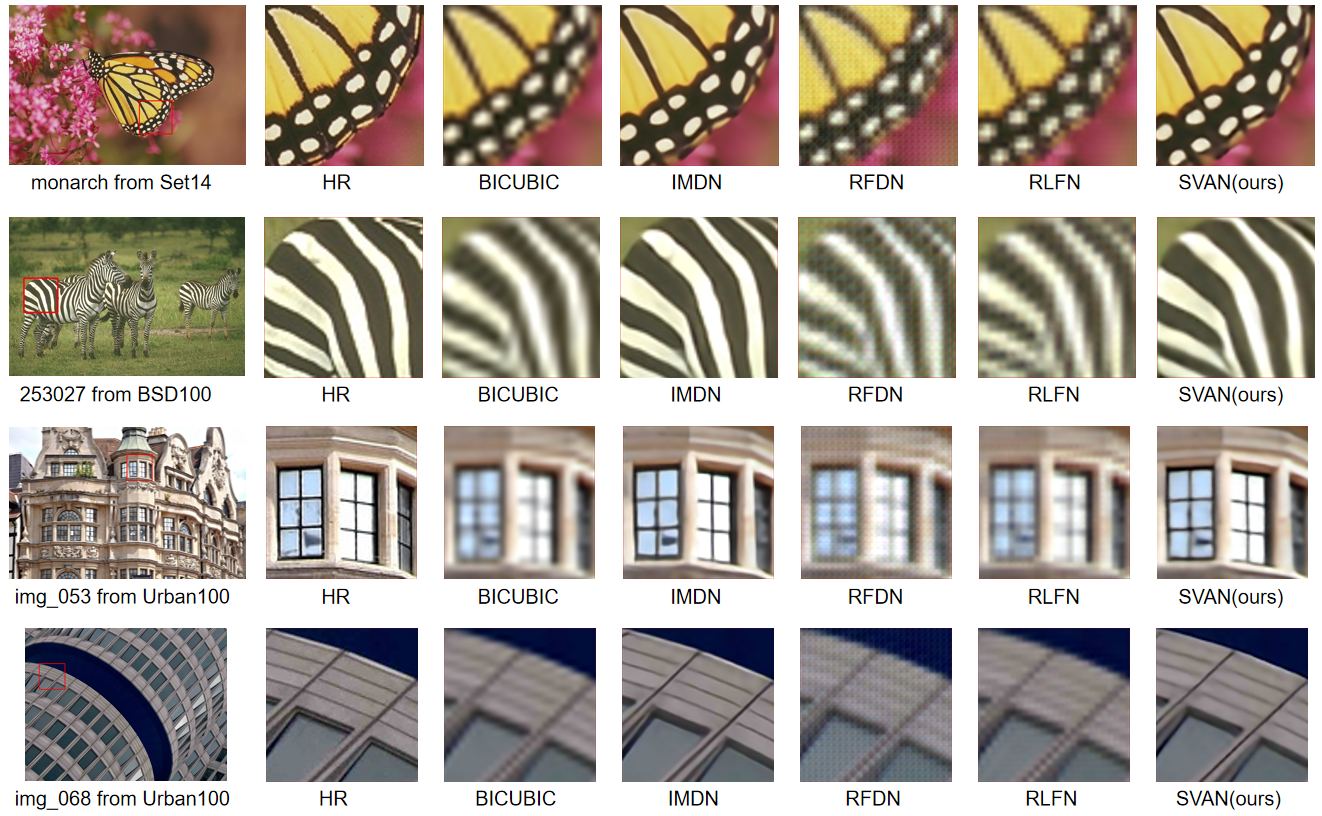}
\end{center}
   \caption{Visual results on benchmark datasets for $\times$4 upscaling. All image comparison results are generated by the code and models provided in the corresponding papers \cite{hui2019lightweight,liu2020residual,kong2022residual}.}
\label{fig:pic}
\end{figure*}

{\bf Qualitative evaluations.} Figure \ref{fig:pic} shows a qualitative comparison of the proposed method on images from Set14, BSD100, and Urban100 at a scale of $\times$4. Although our method's performance on quantitative comparison is slightly lower, it produces images with much better visual qualities. Taking "img\_068" as an example, our method can accurately reconstruct the stripes and line patterns even with a significant reduction in the number of parameters. Most existing methods produce significant artifacts and blurring effects and do not reconstruct the orientation of the stripes correctly. 

\subsection{Ablation study}

{\bf SLKAB's bottleneck and symmetric structures.} We conducted experiments to change the sequence of the attention layer with different receptive field sizes in SLKAB, to verify the rationality of the bottleneck structure and symmetric arrangement of the receptive fields. The experiments were performed on the Set5 dataset with a magnification of 4. The results are shown in Table \ref{table2}.

We order the receptive field sizes in the attentions as 17-1-1-17, 17 means the receptive field size obtained by the combination of 5$\times$5 depth-wise convolution and 5$\times$5 depth-wise dilation convolution with a dilation of 3, and 1 means the receptive field size obtained by 1$\times$1 point convolution. Three other different arrangements were compared 17-1-17-1, 1-17-1-17, and 1-17-17-1. Experiments show that the performance of the 17-1-1-17 receptive field size arrangement is better than other arrangements. The results once again validate that the attention to receptive field bottleneck structure and symmetrical structure can lead to better results in our model.

\begin{table}[ht]
\begin{center}
\scalebox{1}{
\begin{tabular}{|c|c|}
\hline
Structure & Set5\\ \hline
17-1-17-1 & 31.72\\\hline
1-17-1-17 & 31.73\\\hline
1-17-17-1 & 31.74\\\hline
17-1-1-17 & {\bf 31.76}\\
\hline
\end{tabular}}
\end{center}
\caption{Quantitative comparison of the ablation experiment results based on arranging layers with different sizes of receptive fields. {\bf 17} means the large receptive field size obtained by the combination of DW and DW-D convolutions in Figure \ref{fig:slkab}, and {\bf 1} means the small receptive field size obtained by conv-1 in Figure \ref{fig:slkab}. The best results are highlighted in {\bf bold}.}
\label{table2}
\end{table}

\begin{table}[ht]
\begin{center}
\scalebox{0.95}{
\begin{tabular}{|c|c|c|c|}
\hline
Conv & Receptive Field Size & Params[K] & FLOPs[G]\\ \hline
5$\times$5 & 5 & 0.228 & 0.0143\\
17$\times$17 & 17 & 2.604 & 0.1498\\
5-DW \& 5-DW-D & 17 & {\bf 0.156} & {\bf 0.0098}\\
\hline
\end{tabular}}
\end{center}
\caption{Comparison of large kernel convolution combinations with ordinary convolution in terms of efficiency. Calculation on a 256$\times$256 size RGB image. 5-DW corresponds to DW conv which means 5$\times$5 depth-wise convolution in Figure \ref{fig:slkab}, and similarly, 5-DW-D corresponds to DW-D conv which means 5$\times$5 depth-wise dilation convolution in Figure \ref{fig:slkab}. The best results are highlighted in {\bf bold}.}
\label{table3}
\end{table}

{\bf The efficiency of convolution combinations.} We also compare the convolution combination of 5$\times$5 depth-wise convolution and 5$\times$5 depth-wise dilation convolution with a dilation of 3 used in SLKAB with the normal convolution. As shown in Table \ref{table3}, our convolution combination has the same receptive field size as the 17$\times$17 convolution, but the number of parameters is only 6\% of a traditional  17$\times$17 convolution kernel and the FLOPs are also significantly lower. Moreover, the parameters of our large kernel are even smaller than the 5$\times$5 convolution kernel. This comparison indicates that our kernel convolution combination technique is extremely lightweight and efficient. The correctness and practicality of decomposing the large kernel convolution into a combination of depth-wise convolution and depth-wise dilation convolution are again verified.

\section{Conclusion}
\label{Conclusions}
In this paper, we propose a lightweight symmetric visual attention network for efficient SR. Our model uses a combination of different convolutions to greatly reduce the number of parameters while maintaining a large receptive field to ensure reconstruction quality. Then, we form bottleneck attention blocks according to the receptive field size of each layer of convolution and obtain symmetric large kernel attention blocks by symmetric scheduling. The experiment results show that our SVAN achieves efficient SR competitive reconstruction results and reduces the number of parameters by about 70\%. Future work will revolve around improving the quantitative results of SVAN.

\bibliography{svan}

\begin{thebibliography}{42}
\providecommand{\natexlab}[1]{#1}
\providecommand{\url}[1]{\texttt{#1}}
\expandafter\ifx\csname urlstyle\endcsname\relax
  \providecommand{\doi}[1]{doi: #1}\else
  \providecommand{\doi}{doi: \begingroup \urlstyle{rm}\Url}\fi

\bibitem[Agustsson and Timofte(2017)]{agustsson2017ntire}
Eirikur Agustsson and Radu Timofte.
\newblock Ntire 2017 challenge on single image super-resolution: Dataset and
  study.
\newblock In \emph{Proceedings of the IEEE conference on computer vision and
  pattern recognition workshops}, pages 126--135, 2017.

\bibitem[Ahn et~al.(2018)Ahn, Kang, and Sohn]{ahn2018fast}
Namhyuk Ahn, Byungkon Kang, and Kyung-Ah Sohn.
\newblock Fast, accurate, and lightweight super-resolution with cascading
  residual network.
\newblock In \emph{Proceedings of the European conference on computer vision
  (ECCV)}, pages 252--268, 2018.

\bibitem[Bevilacqua et~al.(2012)Bevilacqua, Roumy, Guillemot, and
  Alberi-Morel]{bevilacqua2012low}
Marco Bevilacqua, Aline Roumy, Christine Guillemot, and Marie~Line
  Alberi-Morel.
\newblock Low-complexity single-image super-resolution based on nonnegative
  neighbor embedding.
\newblock In \emph{Proceedings of the British Machine Vision Conference}, 2012.

\bibitem[Chen et~al.(2022)Chen, Wang, Zhou, and Dong]{chen2022activating}
Xiangyu Chen, Xintao Wang, Jiantao Zhou, and Chao Dong.
\newblock Activating more pixels in image super-resolution transformer.
\newblock \emph{arXiv preprint arXiv:2205.04437}, 2022.

\bibitem[Chu et~al.(2021)Chu, Zhang, Ma, Xu, and Li]{chu2021fast}
Xiangxiang Chu, Bo~Zhang, Hailong Ma, Ruijun Xu, and Qingyuan Li.
\newblock Fast, accurate and lightweight super-resolution with neural
  architecture search.
\newblock In \emph{2020 25th International conference on pattern recognition
  (ICPR)}, pages 59--64. IEEE, 2021.

\bibitem[Dai et~al.(2016)Dai, Wang, Chen, and Van~Gool]{dai2016image}
Dengxin Dai, Yujian Wang, Yuhua Chen, and Luc Van~Gool.
\newblock Is image super-resolution helpful for other vision tasks?
\newblock In \emph{2016 IEEE Winter Conference on Applications of Computer
  Vision (WACV)}, pages 1--9. IEEE, 2016.

\bibitem[Ding et~al.(2022)Ding, Zhang, Han, and Ding]{ding2022scaling}
Xiaohan Ding, Xiangyu Zhang, Jungong Han, and Guiguang Ding.
\newblock Scaling up your kernels to 31x31: Revisiting large kernel design in
  cnns.
\newblock In \emph{Proceedings of the IEEE/CVF Conference on Computer Vision
  and Pattern Recognition}, pages 11963--11975, 2022.

\bibitem[Dong et~al.(2016)Dong, Loy, and Tang]{dong2016accelerating}
Chao Dong, Chen~Change Loy, and Xiaoou Tang.
\newblock Accelerating the super-resolution convolutional neural network.
\newblock In \emph{Computer Vision--ECCV 2016: 14th European Conference,
  Amsterdam, The Netherlands, October 11-14, 2016, Proceedings, Part II 14},
  pages 391--407. Springer, 2016.

\bibitem[Dong et~al.(2020)Dong, Wang, Sun, Jia, Gao, and Zhang]{dong2020remote}
Xiaoyu Dong, Longguang Wang, Xu~Sun, Xiuping Jia, Lianru Gao, and Bing Zhang.
\newblock Remote sensing image super-resolution using second-order multi-scale
  networks.
\newblock \emph{IEEE Transactions on Geoscience and Remote Sensing},
  59\penalty0 (4):\penalty0 3473--3485, 2020.

\bibitem[Gao et~al.(2022)Gao, Wang, Li, Li, Yu, and Zeng]{gao2022lightweight}
Guangwei Gao, Zhengxue Wang, Juncheng Li, Wenjie Li, Yi~Yu, and Tieyong Zeng.
\newblock Lightweight bimodal network for single-image super-resolution via
  symmetric cnn and recursive transformer.
\newblock \emph{arXiv preprint arXiv:2204.13286}, 2022.

\bibitem[Gao et~al.(2019)Gao, Zhao, Li, and Tong]{gao2019image}
Qinquan Gao, Yan Zhao, Gen Li, and Tong Tong.
\newblock Image super-resolution using knowledge distillation.
\newblock In \emph{Computer Vision--ACCV 2018: 14th Asian Conference on
  Computer Vision, Perth, Australia, December 2--6, 2018, Revised Selected
  Papers, Part II}, pages 527--541. Springer, 2019.

\bibitem[Gu et~al.(2022)Gu, Kwon, Wang, Ye, Li, Chen, Lai, Chandra, and
  Pan]{gu2022multi}
Jiaqi Gu, Hyoukjun Kwon, Dilin Wang, Wei Ye, Meng Li, Yu-Hsin Chen, Liangzhen
  Lai, Vikas Chandra, and David~Z Pan.
\newblock Multi-scale high-resolution vision transformer for semantic
  segmentation.
\newblock In \emph{Proceedings of the IEEE/CVF Conference on Computer Vision
  and Pattern Recognition}, pages 12094--12103, 2022.

\bibitem[Guo et~al.(2022)Guo, Lu, Liu, Cheng, and Hu]{guo2022visual}
Meng-Hao Guo, Cheng-Ze Lu, Zheng-Ning Liu, Ming-Ming Cheng, and Shi-Min Hu.
\newblock Visual attention network.
\newblock \emph{arXiv preprint arXiv:2202.09741}, 2022.

\bibitem[Guo et~al.(2020)Guo, Luo, He, Huang, and Chen]{guo2020hierarchical}
Yong Guo, Yongsheng Luo, Zhenhao He, Jin Huang, and Jian Chen.
\newblock Hierarchical neural architecture search for single image
  super-resolution.
\newblock \emph{IEEE Signal Processing Letters}, 27:\penalty0 1255--1259, 2020.

\bibitem[Hendrycks and Gimpel(2016)]{hendrycks2016gaussian}
Dan Hendrycks and Kevin Gimpel.
\newblock Gaussian error linear units (gelus).
\newblock \emph{arXiv preprint arXiv:1606.08415}, 2016.

\bibitem[Huang et~al.(2015)Huang, Singh, and Ahuja]{huang2015single}
Jia-Bin Huang, Abhishek Singh, and Narendra Ahuja.
\newblock Single image super-resolution from transformed self-exemplars.
\newblock In \emph{Proceedings of the IEEE conference on computer vision and
  pattern recognition}, pages 5197--5206, 2015.

\bibitem[Hui et~al.(2018)Hui, Wang, and Gao]{hui2018fast}
Zheng Hui, Xiumei Wang, and Xinbo Gao.
\newblock Fast and accurate single image super-resolution via information
  distillation network.
\newblock In \emph{Proceedings of the IEEE conference on computer vision and
  pattern recognition}, pages 723--731, 2018.

\bibitem[Hui et~al.(2019)Hui, Gao, Yang, and Wang]{hui2019lightweight}
Zheng Hui, Xinbo Gao, Yunchu Yang, and Xiumei Wang.
\newblock Lightweight image super-resolution with information
  multi-distillation network.
\newblock In \emph{Proceedings of the 27th acm international conference on
  multimedia}, pages 2024--2032, 2019.

\bibitem[Kim et~al.(2016)Kim, Lee, and Lee]{kim2016deeply}
Jiwon Kim, Jung~Kwon Lee, and Kyoung~Mu Lee.
\newblock Deeply-recursive convolutional network for image super-resolution.
\newblock In \emph{Proceedings of the IEEE conference on computer vision and
  pattern recognition}, pages 1637--1645, 2016.

\bibitem[Kingma and Ba(2014)]{kingma2014adam}
Diederik~P Kingma and Jimmy Ba.
\newblock Adam: A method for stochastic optimization.
\newblock \emph{arXiv preprint arXiv:1412.6980}, 2014.

\bibitem[Kong et~al.(2022)Kong, Li, Liu, Liu, He, Bai, Chen, and
  Fu]{kong2022residual}
Fangyuan Kong, Mingxi Li, Songwei Liu, Ding Liu, Jingwen He, Yang Bai, Fangmin
  Chen, and Lean Fu.
\newblock Residual local feature network for efficient super-resolution.
\newblock In \emph{Proceedings of the IEEE/CVF Conference on Computer Vision
  and Pattern Recognition}, pages 766--776, 2022.

\bibitem[Li et~al.(2021)Li, Sixou, and Peyrin]{li2021review}
Y~Li, Bruno Sixou, and F~Peyrin.
\newblock A review of the deep learning methods for medical images super
  resolution problems.
\newblock \emph{Irbm}, 42\penalty0 (2):\penalty0 120--133, 2021.

\bibitem[Li et~al.(2022)Li, Zhang, Timofte, Van~Gool, Kong, Li, Liu, Du, Liu,
  Zhou, et~al.]{li2022ntire}
Yawei Li, Kai Zhang, Radu Timofte, Luc Van~Gool, Fangyuan Kong, Mingxi Li,
  Songwei Liu, Zongcai Du, Ding Liu, Chenhui Zhou, et~al.
\newblock Ntire 2022 challenge on efficient super-resolution: Methods and
  results.
\newblock In \emph{Proceedings of the IEEE/CVF Conference on Computer Vision
  and Pattern Recognition}, pages 1062--1102, 2022.

\bibitem[Liang et~al.(2021)Liang, Cao, Sun, Zhang, Van~Gool, and
  Timofte]{liang2021swinir}
Jingyun Liang, Jiezhang Cao, Guolei Sun, Kai Zhang, Luc Van~Gool, and Radu
  Timofte.
\newblock Swinir: Image restoration using swin transformer.
\newblock In \emph{Proceedings of the IEEE/CVF international conference on
  computer vision}, pages 1833--1844, 2021.

\bibitem[Lim et~al.(2017)Lim, Son, Kim, Nah, and Mu~Lee]{lim2017enhanced}
Bee Lim, Sanghyun Son, Heewon Kim, Seungjun Nah, and Kyoung Mu~Lee.
\newblock Enhanced deep residual networks for single image super-resolution.
\newblock In \emph{Proceedings of the IEEE conference on computer vision and
  pattern recognition workshops}, pages 136--144, 2017.

\bibitem[Liu et~al.(2020)Liu, Tang, and Wu]{liu2020residual}
Jie Liu, Jie Tang, and Gangshan Wu.
\newblock Residual feature distillation network for lightweight image
  super-resolution.
\newblock In \emph{Computer Vision--ECCV 2020 Workshops: Glasgow, UK, August
  23--28, 2020, Proceedings, Part III 16}, pages 41--55. Springer, 2020.

\bibitem[Liu et~al.(2021)Liu, Lin, Cao, Hu, Wei, Zhang, Lin, and
  Guo]{liu2021swin}
Ze~Liu, Yutong Lin, Yue Cao, Han Hu, Yixuan Wei, Zheng Zhang, Stephen Lin, and
  Baining Guo.
\newblock Swin transformer: Hierarchical vision transformer using shifted
  windows.
\newblock In \emph{Proceedings of the IEEE/CVF international conference on
  computer vision}, pages 10012--10022, 2021.

\bibitem[Liu et~al.(2022)Liu, Mao, Wu, Feichtenhofer, Darrell, and
  Xie]{liu2022convnet}
Zhuang Liu, Hanzi Mao, Chao-Yuan Wu, Christoph Feichtenhofer, Trevor Darrell,
  and Saining Xie.
\newblock A convnet for the 2020s.
\newblock In \emph{Proceedings of the IEEE/CVF Conference on Computer Vision
  and Pattern Recognition}, pages 11976--11986, 2022.

\bibitem[Luo et~al.(2022)Luo, Huang, Yu, Li, Fan, and Liu]{luo2022deep}
Ziwei Luo, Haibin Huang, Lei Yu, Youwei Li, Haoqiang Fan, and Shuaicheng Liu.
\newblock Deep constrained least squares for blind image super-resolution.
\newblock In \emph{Proceedings of the IEEE/CVF Conference on Computer Vision
  and Pattern Recognition}, pages 17642--17652, 2022.

\bibitem[Luo et~al.(2023)Luo, Li, Yu, Wu, Wen, Fan, and Liu]{luo2023fast}
Ziwei Luo, Youwei Li, Lei Yu, Qi~Wu, Zhihong Wen, Haoqiang Fan, and Shuaicheng
  Liu.
\newblock Fast nearest convolution for real-time efficient image
  super-resolution.
\newblock In \emph{Computer Vision--ECCV 2022 Workshops: Tel Aviv, Israel,
  October 23--27, 2022, Proceedings, Part II}, pages 561--572. Springer, 2023.

\bibitem[Martin et~al.(2001)Martin, Fowlkes, Tal, and
  Malik]{martin2001database}
David Martin, Charless Fowlkes, Doron Tal, and Jitendra Malik.
\newblock A database of human segmented natural images and its application to
  evaluating segmentation algorithms and measuring ecological statistics.
\newblock In \emph{Proceedings Eighth IEEE International Conference on Computer
  Vision. ICCV 2001}, volume~2, pages 416--423. IEEE, 2001.

\bibitem[Niu et~al.(2020)Niu, Wen, Ren, Zhang, Yang, Wang, Zhang, Cao, and
  Shen]{niu2020single}
Ben Niu, Weilei Wen, Wenqi Ren, Xiangde Zhang, Lianping Yang, Shuzhen Wang,
  Kaihao Zhang, Xiaochun Cao, and Haifeng Shen.
\newblock Single image super-resolution via a holistic attention network.
\newblock In \emph{Computer Vision--ECCV 2020: 16th European Conference,
  Glasgow, UK, August 23--28, 2020, Proceedings, Part XII 16}, pages 191--207.
  Springer, 2020.

\bibitem[Shamsolmoali et~al.(2019)Shamsolmoali, Li, and
  Wang]{shamsolmoali2019single}
Pourya Shamsolmoali, Xiaofang Li, and Ruili Wang.
\newblock Single image resolution enhancement by efficient dilated densely
  connected residual network.
\newblock \emph{Signal Processing: Image Communication}, 79:\penalty0 13--23,
  2019.

\bibitem[Shi et~al.(2016)Shi, Caballero, Husz{\'a}r, Totz, Aitken, Bishop,
  Rueckert, and Wang]{shi2016real}
Wenzhe Shi, Jose Caballero, Ferenc Husz{\'a}r, Johannes Totz, Andrew~P Aitken,
  Rob Bishop, Daniel Rueckert, and Zehan Wang.
\newblock Real-time single image and video super-resolution using an efficient
  sub-pixel convolutional neural network.
\newblock In \emph{Proceedings of the IEEE conference on computer vision and
  pattern recognition}, pages 1874--1883, 2016.

\bibitem[Sun et~al.(2021)Sun, Gong, Shi, van~den Hengel, and
  Zhang]{sun2021learning}
Wei Sun, Dong Gong, Qinfeng Shi, Anton van~den Hengel, and Yanning Zhang.
\newblock Learning to zoom-in via learning to zoom-out: Real-world
  super-resolution by generating and adapting degradation.
\newblock \emph{IEEE Transactions on Image Processing}, 30:\penalty0
  2947--2962, 2021.

\bibitem[Vaswani et~al.(2017)Vaswani, Shazeer, Parmar, Uszkoreit, Jones, Gomez,
  Kaiser, and Polosukhin]{vaswani2017attention}
Ashish Vaswani, Noam Shazeer, Niki Parmar, Jakob Uszkoreit, Llion Jones,
  Aidan~N Gomez, {\L}ukasz Kaiser, and Illia Polosukhin.
\newblock Attention is all you need.
\newblock \emph{Advances in neural information processing systems}, 30, 2017.

\bibitem[Wang et~al.(2004)Wang, Bovik, Sheikh, and Simoncelli]{wang2004image}
Zhou Wang, Alan~C Bovik, Hamid~R Sheikh, and Eero~P Simoncelli.
\newblock Image quality assessment: from error visibility to structural
  similarity.
\newblock \emph{IEEE transactions on image processing}, 13\penalty0
  (4):\penalty0 600--612, 2004.

\bibitem[Zamir et~al.(2022)Zamir, Arora, Khan, Hayat, Khan, and
  Yang]{zamir2022restormer}
Syed~Waqas Zamir, Aditya Arora, Salman Khan, Munawar Hayat, Fahad~Shahbaz Khan,
  and Ming-Hsuan Yang.
\newblock Restormer: Efficient transformer for high-resolution image
  restoration.
\newblock In \emph{Proceedings of the IEEE/CVF Conference on Computer Vision
  and Pattern Recognition}, pages 5728--5739, 2022.

\bibitem[Zeyde et~al.(2012)Zeyde, Elad, and Protter]{zeyde2012single}
Roman Zeyde, Michael Elad, and Matan Protter.
\newblock On single image scale-up using sparse-representations.
\newblock In \emph{Curves and Surfaces: 7th International Conference, Avignon,
  France, June 24-30, 2010, Revised Selected Papers 7}, pages 711--730.
  Springer, 2012.

\bibitem[Zhang et~al.(2021)Zhang, Zeng, and Zhang]{zhang2021edge}
Xindong Zhang, Hui Zeng, and Lei Zhang.
\newblock Edge-oriented convolution block for real-time super resolution on
  mobile devices.
\newblock In \emph{Proceedings of the 29th ACM International Conference on
  Multimedia}, pages 4034--4043, 2021.

\bibitem[Zhang et~al.(2018)Zhang, Li, Li, Wang, Zhong, and Fu]{zhang2018image}
Yulun Zhang, Kunpeng Li, Kai Li, Lichen Wang, Bineng Zhong, and Yun Fu.
\newblock Image super-resolution using very deep residual channel attention
  networks.
\newblock In \emph{Proceedings of the European conference on computer vision
  (ECCV)}, pages 286--301, 2018.

\bibitem[Zhou et~al.(2020)Zhou, Hou, Chen, Feng, and Yan]{zhou2020rethinking}
Daquan Zhou, Qibin Hou, Yunpeng Chen, Jiashi Feng, and Shuicheng Yan.
\newblock Rethinking bottleneck structure for efficient mobile network design.
\newblock In \emph{Computer Vision--ECCV 2020: 16th European Conference,
  Glasgow, UK, August 23--28, 2020, Proceedings, Part III 16}, pages 680--697.
  Springer, 2020.

\end{thebibliography}
\end{document}